\documentclass{article} 
\usepackage{arxiv}

\usepackage{amsmath,amsfonts,bm}

\def\eqref#1{equation~\ref{#1}}

\def\1{\bm{1}}

\DeclareMathAlphabet{\mathsfit}{\encodingdefault}{\sfdefault}{m}{sl}
\SetMathAlphabet{\mathsfit}{bold}{\encodingdefault}{\sfdefault}{bx}{n}

\usepackage{hyperref}
\usepackage{url}

\usepackage{graphicx}
\usepackage{subcaption}
\usepackage{tabularx}
    \newcolumntype{L}{>{\center\arraybackslash}X}

\usepackage{algorithm}
\usepackage{algorithmic}
\usepackage{enumitem}
\usepackage{booktabs}
\usepackage{wrapfig}

\title{One Timestep is all you need: Training Spiking Neural Networks with Ultra Low Latency}

\author{Sayeed Shafayet Chowdhury, Nitin Rathi \& Kaushik Roy  \\
School of Electrical and Computer Engineering \\
Purdue University\\
West Lafayette, IN 47905, USA \\
\texttt{\{chowdh23,rathi2,kaushik\}@purdue.edu} \\

}

\begin{document}

\maketitle

\begin{abstract}
Spiking Neural Networks (SNNs) are energy efficient alternatives to commonly used deep neural networks (DNNs).~Through event-driven information processing, SNNs can reduce the expensive compute requirements of DNNs considerably, while achieving comparable performance. However, high inference latency is a significant hindrance to the edge deployment of deep SNNs. Computation over multiple timesteps not only increases latency as well as overall energy budget due to higher number of operations, but also incurs memory access overhead of fetching membrane potentials, both of which lessen the energy benefits of SNNs. To overcome this bottleneck and leverage the full potential of SNNs, we propose an Iterative Initialization and Retraining method for SNNs (IIR-SNN) to perform single shot inference in the temporal axis. The method starts with an SNN trained with T timesteps (T$>$1). Then at each stage of latency reduction, the network trained at previous stage with higher timestep is utilized  as initialization for subsequent training with lower timestep. This acts as a compression method, as the network is gradually shrunk in the temporal domain. In this paper, we use direct input encoding and choose T=5, since as per literature, it is  the minimum required latency to achieve satisfactory performance on ImageNet. The proposed scheme allows us to obtain SNNs with up to unit latency, requiring a single forward pass during inference.
We achieve top-1 accuracy of 93.05\%, 70.15\% and 67.71\% on CIFAR-10, CIFAR-100 and ImageNet, respectively using VGG16, with just 1 timestep. In addition, IIR-SNNs perform inference with 5-2500X reduced latency compared to other state-of-the-art SNNs, maintaining comparable or even better accuracy. Furthermore, in comparison with standard DNNs, the proposed IIR-SNNs provide 25-33X higher energy efficiency,  while being comparable to them in classification performance. \footnote{Our code will be made publicly available at \url{ https://github.com/SayeedChowdhury/IIR-SNN}}

\end{abstract}

\section{Introduction}
Deep neural networks (DNNs) have revolutionized the fields of object detection, classification and natural language processing \cite{krizhevsky2012imagenet,hinton2012deep,deng2018deep}, surpassing even human level performance. However, such performance boost has been achieved at the cost of extremely energy-intensive DNN architectures  \cite{li2016evaluating}. Therefore, edge deployment of such DNNs remains a challenge, requiring specialized architectures and training schemes \cite{hadidi2019characterizing}. One approach 
to counter this is adopting model compression
techniques such as pruning \cite{han2015,wen2016}, and data quantization \cite{bnn}. An alternative  route for efficient computation is using Spiking Neural Networks (SNNs) \cite{maass1997networks, roy2019towards}, which are bio-inspired learning frameworks and perform computations using binary spikes instead of analog activations used in standard networks. In this paper, standard networks are referred to as Analog Neural Networks (ANNs), while SNNs operate using spiking inputs.
As such, the sparse event-driven nature of SNNs make them an attractive alternative to ANNs \cite{frenkel2021sparsity}, specially for
resource-constrained machine intelligence.

SNNs derive their energy efficiency primarily from the sparsity of computation at any given timestep and by substituting multiply–accumulate (MAC) operations in ANNs with additions. However, ANNs and current SNNs differ in terms of number of forward passes required for each input; ANNs infer in a single shot, whereas SNNs require computation over multiple timesteps, resulting in high inference latency. The requirement of multi-timestep processing causes threefold challenges in SNNs in terms of edge deployment- (i) too long latency might be unsuitable for real-time applications, (ii) accumulation of spikes over timesteps results in higher number of operations, thereby reducing efficiency, and (iii) accumulated neuronal membrane potentials of previous steps are used for computation at each timestep which adds the requirement of additional memory to store the intermediate potentials as well as the expensive memory access cost to the total power budget. As a result, reduction of inference latency while maintaining performance is a critical research problem in SNNs. Again, surrogate-gradient based SNN training with multiple timesteps is performed using backpropagation through time (BPTT) \cite{neftci2019surrogate}, similar to Recurrent Neural Networks (RNNs), with membrane potential holding the internal state having recurrent connections. However, DNN topologies indicate that for static image classification tasks, recurrent connections are not required, rather feed-forward networks are enough to reach convergence. This motivates us to explore the possibility of obtaining SNNs where inference is performed with just one forward pass.

Currently, most of the commonly used SNNs use Poisson rate coding \cite{diehl2015fast,sengupta2019going}, where the firing rate of the spiking neuron approximates the analog activation of corresponding ANN unit. This scheme provides comparable results to full precision ANNs, however it suffers from high inference latency \cite{sengupta2019going,rmp}. In rate-coding, the discretization error caused by representing analog values with spike-train is inversely proportional to the number of timesteps. Hence, a large number of timesteps is usually adopted to obtain high performance \cite{sengupta2019going}. As such, it is a challenge to reduce latency considerably without sacrificing performance degradation. On the other hand, with multiple timestep computation, the total energy consumption increases significantly due to the costly memory access operations for fetching membrane potentials. Though most of the previous SNN works ignore it and consider the computational cost only, the memory access cost can in fact be orders of magnitude higher compared to floating point arithmetic operations \cite{deepcompresion}. This bottleneck can be overcome if SNNs can be trained to infer in one timestep, no longer requiring the excess memory access operations due to membrane potential. To that effect, our goal here is to reduce latency all the way up to unity (lowest limit) without hampering accuracy significantly. 

In this paper, we propose an iterative training scheme to reduce the latency requirement of SNNs. We adopt the hybrid method \cite{rathi2020enabling}, so, first an ANN is trained, followed by ANN-SNN conversion and SNN domain training with backpropagation. The scheme utilizes direct input encoding \cite{rathi2020diet,zheng2021going}, since it has enabled training SNNs with 5-6 timesteps only. The threshold of neurons at each layer is a learnable parameter in our method. Starting with an SNN trained with 5 timestep, we gradually reduce latency by iterative retraining. At  each timestep  reduction stage,  the  network  trained  at previous  iteration  with  higher  timestep  is  utilized  as  initialization  for  subsequent training with lower timestep. Direct transition from ANN or 5 timestep SNN to 1 timestep for deep SNNs results in training failures due to spike vanishing at the deeper layers. However, the proposed iterative process enables some spike propagation till the final layer through which backpropagation can start training and proper layerwise thresholds can be learnt for each corresponding timestep. This method gradually compresses the temporal axis of SNNs, since reduction of each timestep  removes one forward pass from the computational graph. 

To summarize, the main contributions of this work are as follows,
\vspace{-.5em}

\begin{itemize}[leftmargin=*]
\setlength{\itemindent}{0em}

\item We present an iterative training technique to obtain deep SNNs which infer using a single timestep, which to the best of our knowledge, is the first SNN method to achieve competitive classification performance on ImageNet using unit timestep.

\item Since inference is done in one timestep (single forward pass), our approach does not require the memory access cost of membrane potentials, unlike previously proposed SNNs.

\item The proposed IIR-SNNs are able to infer with 5-2500X lower timesteps compared to other state-of-the-art SNNs, while achieving comparable or even better accuracy.

\item 1 timestep inference 
also provides ultra-efficient computation due to low spike rate, resulting in SNNs which are 25-33X energy efficient compared to iso-architecture and iso-performing ANNs. 

\item The proposed sequential training scheme enables obtaining deep-Q network based reinforcement learning agents using SNN on Cartpole and Atari pong environments which infer using 1 timestep. 

\end{itemize}

\vspace{-5mm}
\section{Related Works}
\label{related_works}

\textbf{ANN-SNN Conversion.} A widely used approach of obtaining deep SNNs involves training an ANN 
and converting to SNN for finetuning \cite{cao2015spiking, diehl2015fast,sengupta2019going}. To minimize conversion loss, the ANNs are usually trained without bias, batch-norm or maxpooling layers, though some works bypass these constraints using custom layers  \cite{rueckauer2017conversion}. Proper layerwise threshold adjustment is critical to convert ANNs to SNNs successfully. One approach of threshold balancing is to choose the layerwise thresholds as the maximum pre-activation of the neurons \cite{sengupta2019going}. While this provides high accuracy, the associated drawback is high inference latency (about 1000 timesteps). Alternatively, \cite{rueckauer2017conversion} suggest choosing a certain percentile of the pre-activation distribution as the threshold to reduce inference latency and improve robustness. However, these conversion based methods \cite{rueckauer2017conversion,rmp} still require few hundred timesteps to perform satisfactorily.

\textbf{Backpropagation from Scratch and Hybrid Training.} An alternate route of training SNNs with reduced latency is learning from scratch using backpropagation (bp). To circumvent the challenge posed by non-differentiability of the spike function, surrogate gradient based optimization 
has been proposed \cite{neftci2019surrogate} to implement bp in SNNs effectively \cite{huh2018gradient,lee2020enabling}. \cite{zenke2018superspike} propose surrogate gradient based bp on the membrane potential, while
\cite{shrestha2018slayer} perform bp based on the gradients calculated using the difference between the membrane potential and the threshold, but results seem limited to MNIST using shallow architectures. \cite{wu2018spatio} also perform backpropagation through time (BPTT) on SNNs with a surrogate gradient defined on the continuous-valued membrane potential. Overall, surrogate-gradients based BPTT training has resulted in SNNs with high accuracy, but the training is very compute
intensive compared to conversion techniques and the latency is still significant ($\sim$100-125 timesteps).
\cite{rathi2020enabling} propose a hybrid approach where a pre-trained ANN is used as initialization for subsequent surrogate gradient SNN learning. This hybrid approach improves upon conversion by reducing latency and speeds up convergence of direct bp from scratch method. 

\textbf{Temporal Encoding.} A different approach to train SNNs is based on the timing of spike of neurons. Temporal coding schemes such as phase \cite{kim2018deep} or burst \cite{park2019fast} coding attempt to capture temporal information into the spike trains, but incur high latency and large number of spikes. A related method is time-to-first-spike (TTFS) coding \cite{zhang2019tdsnn,park2020t2fsnn}, where each neuron is allowed to spike just once but the high latency persists.
\cite{comsa2020temporal} propose using relative timing of spikes to encode information, but  only report results for simple tasks like MNIST and its scalability to deeper architectures and more complex datasets remains unclear. Notably, since our proposed scheme enables one step inference, the neurons are automatically limited to maximum one spike per neuron, however our latency is significantly lower.

\textbf{Direct Encoding.} The analog pixels are directly applied to the 1$^{st}$ convolutional layer in direct encoding \cite{rueckauer2017conversion,lu2020exploring, rathi2020diet,zheng2021going}. Using this method, \cite{rathi2020diet} achieve competitive performance on ImageNet using just 5 timesteps. Again, threshold-dependent batch normalization is employed together with direct encoding by \cite{zheng2021going} to obtain high performing SNNs with 6 timesteps only. Inspired by such promising performance, we adopt the direct encoding method in this work. 

\section{Background}

\textbf{Spiking Neuron Model}
The Leaky Integrate and Fire (LIF) SNN model \cite{hunsberger2015spiking} is described as-
\begin{equation}\label{eqn:1}
\tau_{m} \frac{dU}{dt}= -(U-U_{rest}) + RI,~~~~U\leq V_{th}
\end{equation}
where $U$, $I$, $\tau_{m}$, $R$, $V_{th}$ and $U_{rest}$ denote membrane potential, input  representing weighted sum of spikes, time constant for membrane potential decay,  leakage path resistance,  firing threshold and resting potential, respectively. We employ a discretized version of Eqn.~\ref{eqn:1} given as-
\begin{equation}\label{eqn:2}
u_i^t=\lambda_i u_i^{t-1}+\sum_{j} w_{ij}o_j^t-v_{i}o_i^{t-1},
\end{equation}
\begin{equation}\label{eqn:3}
z_i^{t-1}=\frac{u_i^{t-1}}{v_i}~~~\text{and}~~~o_i^{t-1}=\left\{
                \begin{array}{ll}
                  1,~~~\text{if}~u_i^{t-1}>v_{i}\\
                  0,~~~\text{otherwise}
                \end{array}
              \right.
\end{equation}
where $u$ is the membrane potential, subscripts $i$ and $j$ represent the post and pre-neuron, respectively, t denotes timestep, $\lambda$ is the leak constant= $e^{\frac{-1}{\tau_{m}}}$, $w_{ij}$ represents the weight between the i-th and j-th neurons, $o$ is the output spike, and $v$ is the
threshold. As indicated by Eqn.~\ref{eqn:2}, we implement a soft-reset, so, $u$ is reduced by $v$ upon crossing the threshold. The details of training SNN with a certain inference timestep is provided in appendix section \ref{app1}. In particular, Algorithm \ref{alg:surrogate} depicts the training scheme for one iteration. We also train the thresholds and leaks using bp; 
the implementation is publicly available at \url{ https://github.com/SayeedChowdhury/IIR-SNN}.

\begin{figure}[t]
\vspace{-8mm}
  \centering
   \includegraphics[width=\linewidth]{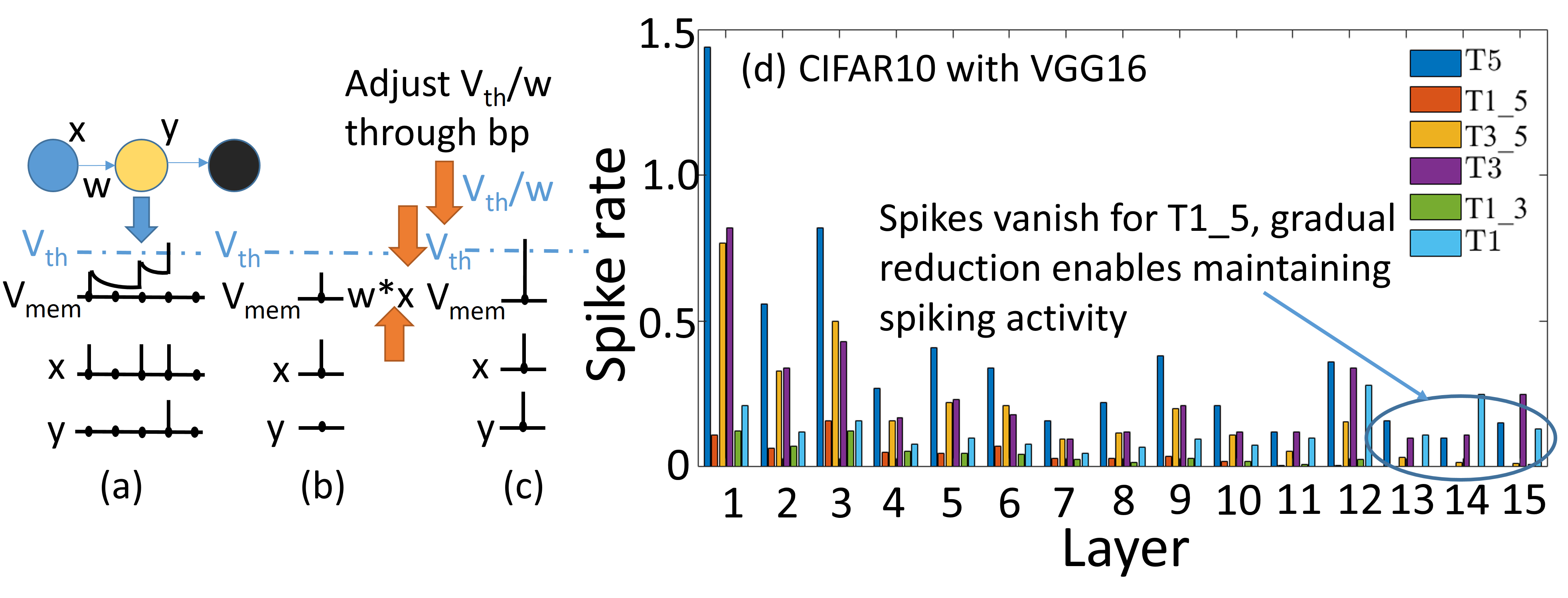}
  
   \caption{(a) Schematic of an SNN with dynamics shown for yellow neuron with x as input and y as output, (b) No output spike for direct transition from 5 to 1 timestep, (c) Output spike in just 1 timestep through $V_{th}/w$ lowering, (d) Layerwise spike rates, Tx represents a converged SNN with `x' timestep, Tx\textunderscore y represents an SNN with `x' timestep initialized with a `y' timestep trained SNN.}
  \label{fig:fig1}
\vspace{-1.3em}
\end{figure}

\section{Proposed Latency Reduction Method}

The starting point of our training pipeline follows the proposed methodology in \cite{rathi2020diet}, difference being we utilize batch-norm (bn) during ANN training and subsequently the bn parameters are fused with the layerwise weights as done in \cite{rueckauer2017conversion}. With this pretrained ANN, the weights are copied to an iso-architecture SNN and we select the 90.0 percentile of the pre-activation distribution at each layer as its threshold. Then the SNN is trained for 5 timesteps using bp which serves as our baseline; training steps are detailed in Algorithm \ref{alg:surrogate}. Our starting point is chosen as a  5 timesteps trained network, since it is lowest latency that previous works have reported for SNN ImageNet training \cite{rathi2020diet,zheng2021going} with high performance. Going below 5 results in convergence failure due to spike vanishing at deeper layers.

\textbf{Direct Inference with Unit Timestep.} As mentioned in Section 1, our goal is to obtain SNNs with unit latency to harness the maximum efficiency possible. To that effect, next we explore the feasibility of directly reducing latency from 5 to 1. Fig.~1(a) schematically depicts an SNN with 3 neurons (one per layer), we focus on the yellow neuron. Suppose it receives x as input and y is its output. With the weight (w) and threshold (V$_{\text{th}}$) trained for 5 timesteps, there is enough accumulation of membrane potential (V$_{\text{mem}}$) to cross V$_{\text{th}}$ and propagate spikes to next layer within that 5 timestep window. However, when we try to perform inference with 1 timestep (Fig.~1(b)), there is no output spike as the V$_{\text{mem}}$ is unable to reach V$_{\text{th}}$ instantly. So, the w and V$_{\text{th}}$ need to be adjusted such that information can propagate even within 1 step. However, balancing the thresholds properly is critical for SNNs to perform well. If V$_{\text{th}}$ is too high, spikes vanish, but lowering V$_{\text{th}}$ too much results in too many spikes, leading to unsatisfactory performance \cite{zheng2021going}. Hence, our goal is adjusting V$_{\text{th}}$/w through learning using bp, so that only neurons salient for information propagation are able to spike in 1 timestep (Fig.~1(c)), while other neurons remain dormant. 

\vspace{-1mm}
\textbf{Direct Transition to Training with Unit Timestep.} Next, we begin training with 1 timestep initialized with the 5 timestep trained SNN, however, the network fails to train. To investigate this, we plot the layerwise spike rates for a VGG16 on CIFAR10 in Fig.~1(d). For the rest of this paper, with Tx and Tx\textunderscore y, we denote converged SNN with `x' timestep and an SNN with `x' timestep initialized with a `y' timestep trained SNN, respectively. While T5 has sufficient spiking activity till final layer, for T1\textunderscore 5, spikes die out in the earlier layers. Due to such spike vanishing, all outputs at the end layer are 0, thus bp cannot start training. Again, this happens since the initial V$_{\text{th}}$ and w are trained for 5 timesteps, so the drastic transition to 1 timestep hinders spikes to propagate till the end.

\vspace{-1mm}
\textbf{Gradual Latency Reduction.} To mitigate the above issue, we adopt a gradual latency reduction approach. We observe that for T3\textunderscore 5, though layerwise spike activity decays significantly, there is still some spikes that reach the final layer. So, we start training with 3 timesteps using T5 as initialization and training is able to converge through bp. The spiking activity is recovered when learning converges as shown in Fig.~1(d), case T3. Subsequently, we train a network with just 1 timestep by initializing it with T3, and successfully attain convergence. The results for this case is shown as T1 in Fig.~1(d). Motivated by these observations, we propose an Iterative Initialization and Retraining method for training SNNs with lowest possible latency (1 timestep), abbreviated as IIR-SNN.
\begin{wrapfigure}{R}{0.5\textwidth}
    \begin{minipage}{0.5\textwidth}
    \vspace{-4mm} 
      \begin{algorithm}[H]
   \caption{Pseudo-code for IIR-SNN training.}
   \label{alg:iirsnn}
\begin{algorithmic}
   \STATE {\bfseries Input:}  Trained SNN with $N$ timesteps ($TN$), timesteps reduced per reduction step ($x$), number~of epochs to train ($e$)
   \STATE {\bfseries Initialize:} new SNN initialized with trained parameters of $TN$, reduced   latency $T_r = N-x$
   
   \WHILE{$T_{r}>0$}
   
   \STATE //~{\bfseries Training Phase}
   \FOR{$epoch \leftarrow 1$ {\bfseries to} $e$}
          
        \STATE //Train network with $T_r$ timesteps using algorithm \ref{alg:surrogate}

   \ENDFOR
   \STATE //~ Initialize another iso-architecture SNN with parameters of above trained network
   \STATE //~ Timestep reduction
   \STATE $T_r = T_r-x$
   \ENDWHILE
   
   \end{algorithmic}
\vspace{-1mm} 
\end{algorithm}
    \end{minipage}
    \vspace{-5mm} 
  \end{wrapfigure}
A pseudo-code of IIR-SNN training is given in Algorithm \ref{alg:iirsnn}. Beginning with T5, we gradually reduce the latency by 1 at each step and train till convergence. While training with a certain latency, the network is initialized with the higher timestep network trained at previous training stage.

\textbf{Latency Reduction as Form of Compression.} SNNs with multiple timesteps are trained with BPTT \cite{neftci2019surrogate}, like RNNs. If we unroll SNNs in time, it becomes obvious that each timestep adds a hidden state to a spiking neuron. Thus, reducing latency compresses the SNNs in the temporal axis and the eventual unit timestep network consists of a single forward pass, with no internal states. From a related perspective, we can perceive gradual latency reduction as sequential temporal distillation, where at each step, the network with higher timesteps acts as the teacher, while the student network with reduced latency learns from it. Similar sequential distillation (training by generations) has been implemented in ANNs \cite{furlanello2018born,yang2019training} to obtain better performing students. However, in our case, there is no explicit distillation loss and the sequential training is a requirement for convergence instead of being just a performance enhancement tool. Also, unlike ANN sequential distillation, the student architecture remains same spatially throughout all generations in the proposed method and compression occurs in the temporal domain of SNNs.
\vspace{-3mm}
\section{Experiments and Results}
\label{results}
\vspace{-3mm}
\textbf{Datasets and Models.}
We perform experiments on CIFAR10, CIFAR100  and ImageNet using VGG16 and ResNet20 architectures. Some small scale studies involve VGG6. We also investigate the performance of the proposed IIR-SNN in reinforcement learning (RL) applications using Cartpole and Atari-pong. Appendix \ref{app2} includes 
architectural details and training hyperparameters. 

\vspace{-1mm}
\textbf{Results on CIFAR and ImageNet.} The experimental results using the proposed IIR-SNN scheme for CIFAR and ImageNet datasets are shown in Table \ref{acc}. We achieve top-1 accuracy of 93.05\% and 70.15\% on CIFAR-10 and CIFAR-100, respectively using VGG16 architecture, with just 1 timestep (T1); results using ResNet20 are shown too. Next, to investigate the scalability of the proposed algorithm, we experiment with ImageNet where we obtain 67.71\% top-1 accuracy with T1. As we sequentially reduce the timestep from 5 to 1, there is a slight accuracy degradation, however that is due to the inherent accuracy versus latency trade-off in SNNs. Notably, the proposed scheme allows us to bring the SNN latency down to lowest possible limit (1 timestep) while maintaining comparable accuracy to corresponding ANN as well as the 5 timestep SNN. Additionally, this provides the option to choose the SNN model suitable for application in hand from a range of timesteps; if latency and efficiency is the priority, T1 would be preferred, albeit with slightly lower accuracy compared to T5.

\begin{table}[t]
\vspace{-10mm}
\centering
\caption{Top-1 classification accuracy (\%) using IIR-SNN, Tx denotes SNN with `x' timestep}
\label{acc}
\renewcommand{\arraystretch}{1.15}

\setlength{\tabcolsep}{15pt}

       \begin{tabularx}{\textwidth}{cccccccc}
       
\hline

Architecture & Dataset & ANN & T5 & T4 &T3 &T2 &T1 \\
\hline

VGG6& CIFAR10 & 91.59 & 90.61 & 90.52 & 90.40 & 90.05& 89.10 \\
\hline

VGG16& CIFAR10 & 94.10 & 93.90 & 93.87 & 93.85 & 93.72& 93.05 \\
\hline

ResNet20& CIFAR10 & 93.34 & 92.62 & 92.58 & 92.56 & 92.11& 91.10 \\
\hline

VGG16& CIFAR100 & 72.46&71.58& 71.51& 71.46 &71.43&70.15 \\
\hline

ResNet20& CIFAR100 & 65.90 & 65.57 & 65.37 & 65.16 & 64.86& 63.30\\
\hline

VGG16& ImageNet &  70.08&69.05&69.03&69.01&68.62&67.71\\
\hline
\vspace{-10mm}
\end{tabularx}
\end{table}
\vspace{-1mm}
\textbf{Performance Comparison.} Next, we  compare our performance with  different  state-of-the-art  SNNs  in Table \ref{comparison}. IIR-SNN performs better than or comparably to all these methods, while achieving significantly lower inference latency. In particular, note that previously it was challenging to obtain satisfactory performance with low latency on ImageNet, with lowest reported latencies of 5 \cite{rathi2020diet} and 6 \cite{zheng2021going}. In contrast, we report 67.71\% top-1 accuracy on ImageNet using T1. Again, \cite{wu2019direct} achieve 90.53\% accuracy on CIFAR-10 using 12 timesteps with direct input encoding (similar to IIR-SNN), but the proposed method enables us to obtain 93.05\% accuracy on CIFAR-10 using T1. For CIFAR100, \cite{rathi2020diet} report 69.67\% accuracy with 5 timesteps, whereas IIR-SNN achieves 70.15\% with T1. Overall, IIR-SNN demonstrates 5-2500X improvement in inference latency compared to
other works while maintaining iso or better classification performance. Table \ref{comparison} also demonstrates the continuum of training SNNs from ANN-SNN conversion to T1 IIR-SNN. Initial ANN-SNN conversion methods required latency on the order of thousands \cite{sengupta2019going,hunsberger2015spiking}. Surrogate-gradient based bp  \cite{wu2019direct,lee2020enabling} reduced it to few tens to hundred, but scaling up to ImageNet was challenging. Hybrid training \cite{rathi2020enabling} combined these two methods to bring the latency down to few hundreds. Subsequently, direct input encoding enabled convergence on ImageNet with latency of $\sim5$ \cite{rathi2020diet,zheng2021going}. The proposed IIR-SNN method leverages all these previously proposed techniques and improves upon them by incorporating the sequential reduction approach to obtain unit latency. We also achieve better performance compared to different SNN encoding schemes such as TTFS \cite{park2020t2fsnn}, phase \cite{kim2018deep}, burst \cite{park2019fast}; detailed comparison is provided in Appendix section \ref{app3-1}.

\begin{table}[t]
\vspace{-11mm}
\small
\centering
\caption{Comparison of IIR-SNN to other reported results. SGB, hybrid and TTFS denote surrogate-gradient based backprop, pretrained ANN followed by SNN fine-tuning, and time-to-first-spike scheme, respectively  and (qC, dL) denotes an architecture with q conv layers and d linear layers.}
\label{comparison}
\renewcommand{\arraystretch}{1.15}

\setlength{\tabcolsep}{15pt}

       \begin{tabularx}{\textwidth}{cccccc}
       
\hline

Reference & Dataset & Training & Architecture & Accuracy(\%) & Timesteps \\
\hline

\cite{hunsberger2015spiking}  &CIFAR10 & 
Conversion        & 2C,
2L  &82.95  & 6000\\ 
\hline

\cite{cao2015spiking}  &CIFAR10 & 
Conversion        & 3C,
2L  &77.43  & 400\\
\hline

\cite{sengupta2019going}  &CIFAR10 & 
Conversion        & VGG16  &91.55  & 2500\\
\hline

\cite{lee2020enabling}  &CIFAR10 & SGB        & VGG9  &90.45  & 100\\
\hline

\cite{rueckauer2017conversion}  &CIFAR10 & 
Conversion        & 4C,
2L  &90.85  & 400\\
\hline
\cite{rathi2020enabling}  &CIFAR10 & Hybrid        & VGG9  &90.5  & 100\\
\hline

\cite{park2020t2fsnn}  &CIFAR10 & TTFS        &  VGG16 &91.4  & 680\\
\hline
\cite{park2019fast}  &CIFAR10 & 
Burst-coding        & VGG16  &91.4  & 1125\\
\hline
\cite{kim2018deep}  &CIFAR10 & Phase-coding        & VGG16  &91.2  & 1500\\
\hline

\cite{wu2018spatio}  &CIFAR10 & SGB        & 2C, 2L  & 50.7 & 30\\
\hline

\cite{wu2019direct}  &CIFAR10 & SG      & 5C, 2L &90.53  & 12\\
\hline 

\cite{wu2019tandem} &CIFAR10 &Tandem Learning& 5C, 2L  &90.98  & 8\\
\hline

\cite{zhang2020temporal} &CIFAR10 &SGB& 5C, 2L  &91.41  & 5\\
\hline

\cite{rathi2020diet} &CIFAR10 &Hybrid& VGG16  &92.70  & 5\\
\hline

\cite{zheng2021going} &CIFAR10 &STBP-tdBN& ResNet-19  &93.16  & 6\\
\hline

\bf This work  &\bf CIFAR10 & \bf IIR-SNN        & \bf VGG16  &\bf 93.05  & \bf1\\
\hline
\cite{lu2020exploring}  &CIFAR100 & 
Conversion        & VGG15  &63.2  & 62\\
\hline
\cite{rathi2020enabling}  &CIFAR100 & Hybrid        & VGG11  &67.9  & 125\\
\hline

\cite{park2020t2fsnn}  &CIFAR100 & TTFS       &  VGG16 &68.8  & 680\\
\hline
\cite{park2019fast}  &CIFAR100 & 
Burst-coding        & VGG16  &68.77  & 3100\\
\hline
\cite{kim2018deep}  &CIFAR100 & Phase-coding        & VGG16  &68.6  & 8950\\
\hline
\cite{rathi2020diet} &CIFAR100 &Hybrid& VGG16  &69.67  & 5\\
\hline

\bf This work  &\bf CIFAR100 & \bf IIR-SNN        & \bf VGG16  &\bf 70.15  & \bf1\\
\hline

\cite{sengupta2019going}  &Imagenet & 
Conversion        & VGG16  &69.96 & 2500\\
\hline
\cite{rueckauer2017conversion}  &Imagenet & 
Conversion        & VGG16 &49.61  & 400\\
\hline
\cite{rathi2020enabling}  &Imagenet & Hybrid        & VGG16  &65.19  & 250\\
\hline
\cite{wu2019tandem} &Imagenet &Tandem Learning& AlexNet  &50.22  & 10\\
\hline

\cite{lu2020exploring}  &Imagenet & 
Conversion        & VGG15  &66.56  & 64\\
\hline
\cite{rathi2020diet} &Imagenet&Hybrid& VGG16  &69.00  & 5\\
\hline
\cite{zheng2021going} &Imagenet &STBP-tdBN& ResNet-34  & 67.05  & 6\\
\hline

\bf This work  &\bf Imagenet & \bf IIR-SNN        & \bf VGG16 &\bf \bf 67.71  & \bf 1\\
\hline
\vspace{-10.5mm}
\end{tabularx}
\end{table}

\begin{figure}[t]
\vspace{-8mm}
  \centering
   \includegraphics[width=\linewidth]{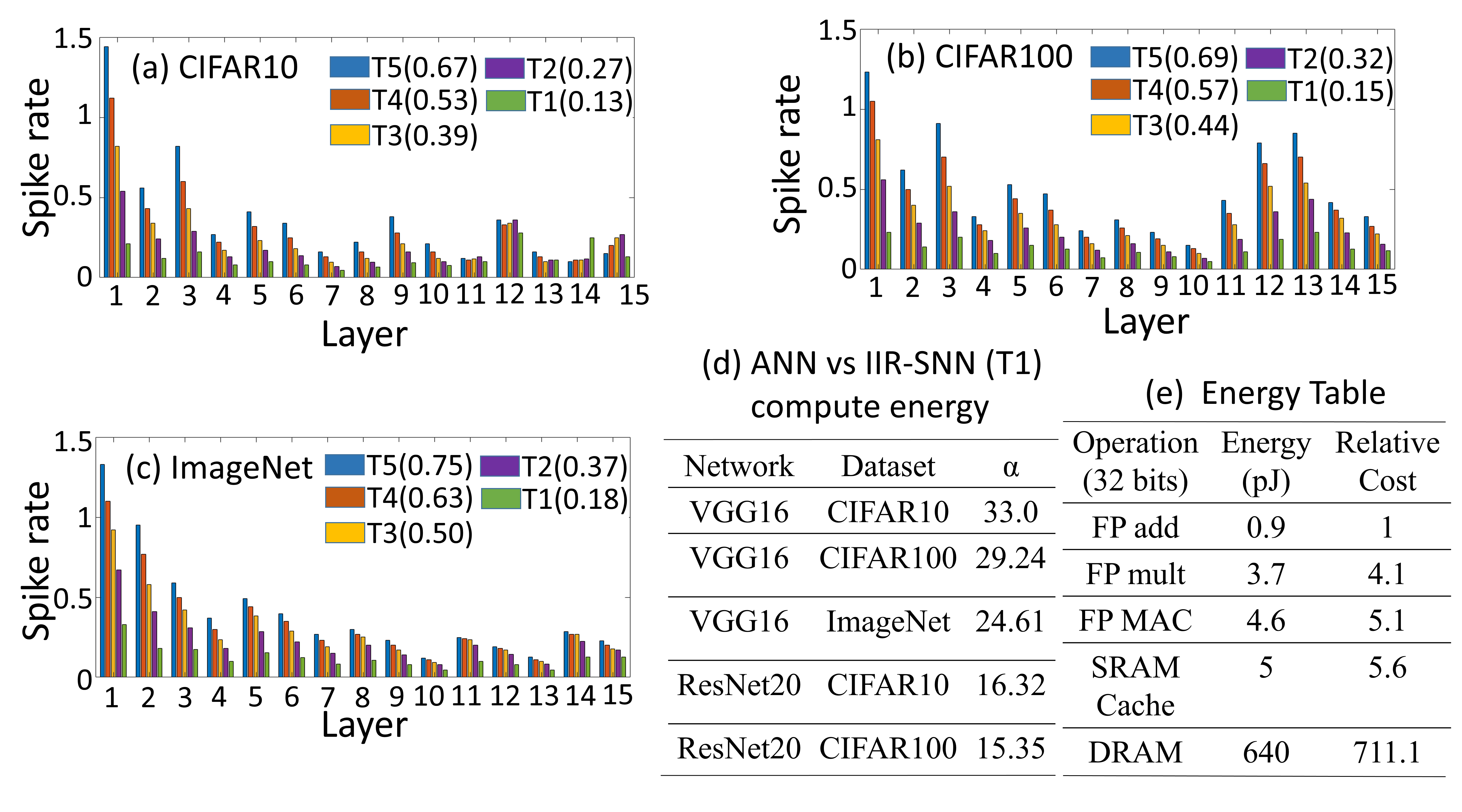}
  \vspace{-2.5em}
   \caption{Layerwise spike rates for a VGG16 across T5 to T1 (average spike rate in parenthesis) on (a) CIFAR10, (b) CIFAR100, (c) ImageNet, (d) relative cost of compute between ANN and T1, (e) operation-wise energy consumption in 45nm CMOS technology  \cite{horowitz20141}.}
  \label{fig:fig2}
\vspace{-1.5em}
\end{figure}

\vspace{-1.5mm}
\textbf{Inference Efficiency.}  Next, we compare the energy efficiency of IIR-SNNs with ANNs and multi-timestep SNNs. In SNNs, the floating-point (FP) additions replace the FP MAC operations. This results in higher compute efficiency as the cost of a MAC ($4.6pJ$) is $5.1\times$ to an addition ($0.9pJ$) \cite{horowitz20141} in 45nm
CMOS technology (as shown in Fig.~\ref{fig:fig2}(e)). Appendix \ref{app3-2} contains the equations of computational cost in the form of operations per layer in an ANN, $\#\text{ANN}_{\text{ops}}$. For IIR-SNN, the number of operations is given as $\#\text{IIR-SNN}_{\text{ops, q}} = \text{spike rate}_\text{q} \times \#\text{ANN}_{\text{ops, q}};$ spike rate$_\text{q}$ denoting the average number of spikes per neuron per inference over all timesteps in layer q. The layerwise spike rates across T5 to T1 are shown in Fig.~\ref{fig:fig2}(a-c). Note, the spike rates decrease significantly with latency reduction from 5 to 1, leading to considerably lower operation count in T1 compared to T5. The overall average spike rates using T1 for CIFAR-10, CIFAR-100 and ImageNet on VGG16 are 0.13, 0.15 and 0.18, respectively; all significantly below 5.1 (relative cost of MAC to addition), indicating the energy benefits of IIR-SNN over the corresponding ANN. The first layer with direct input encoded SNNs receive analog inputs, hence the operations are same as an ANN at this layer. Considering it,
we compute the compute energy benefits of IIR-SNN over ANN, $\alpha$ as , 
\vspace{-2mm}
\begin{equation}
    \alpha = \frac{\text{E}_{\text{ANN}}}{\text{E}_{\text{IIR-SNN}}}=
    \frac{\sum_{q=1}^{L} \#\text{ANN}_{\text{ops,q}}*4.6}{\#\text{IIR-SNN}_{\text{ops,1}}*4.6+\sum_{q=2}^{L} \#\text{IIR-SNN}_{\text{ops,q}}*0.9}.
    \vspace{-2mm}
\end{equation}
 The values of 
 $\alpha$ for different datasets and architectures are given in Fig.~\ref{fig:fig2}(d). For VGG16, we obtain $\alpha$ of 33.0, 29.24 and 24.61 on CIFAR-10, CIFAR-100 and ImageNet, respectively. 
  Besides compute energy, another significant overhead in SNNs occurs due to memory access costs which can be 2-3 orders of magnitude higher \cite{deepcompresion} compared to FP adds as shown in Fig.~\ref{fig:fig2}(e). In multi-timestep SNNs, weights and membrane potentials have to be fetched at every timestep, incurring significant data movement cost. However, most previous works \cite{park2020t2fsnn, rathi2020diet, zheng2021going, rathi2020enabling} did not consider this cost while comparing the energy benefits of SNN to ANN. To obtain a fairer comparison, we analyze the costs taking the memory access issue into consideration.  For multi-timestep SNNs, in addition to the weights, V$_{\text{mem}}$ needs to be stored and fetched at each timestep for all neurons, unlike ANNs. However, the memory requirements for T1 SNNs are same as ANNs; T1 SNNs neither require any extra memory to store V$_{\text{mem}}$ for future timesteps, nor do they have any extra memory accesses compared to ANNs as inference is single shot. The actual improvements in energy due to this depends on the hardware architecture and system configurations. Hence, we compare the reduction in terms of number of memory access. For a VGG16, the proposed T1 reduces the number of memory accesses by 5.03$\times$ compared to \cite{rathi2020diet}. More generally, our scheme reduces the number of memory accesses by approximately T$\times$ compared to an SNN trained with T timesteps.
\vspace{-3mm}

\begin{wraptable}{r}{0.35\textwidth}
\vspace{-4mm}
\small
\caption{Comparison of IIR-SNN to ANNb}\label{annb}
\vspace{-5mm}
\setlength{\tabcolsep}{7pt}
\begin{tabular}{ccc}\\\toprule  
Reference & Dataset & Accuracy(\%)\\\midrule
\cite{sakr2018true} &CIFAR10 & 89.6\\  \midrule
\cite{bengio2013estimating} &CIFAR10 & 85.2\\  \midrule
\cite{optimal}&CIFAR10 & 91.88\\  \bottomrule
IIR-SNN&CIFAR10 & 93.05\\  \bottomrule
\cite{optimal}&CIFAR100 & 70.43\\  \bottomrule
IIR-SNN&CIFAR100 & 70.15\\  \bottomrule
\end{tabular}
\vspace{-3mm}
\end{wraptable} 
\vspace{1.5mm}
\textbf{Comparison with binary activated ANNs.} A key distinction between ANNs and SNNs is the notion of time, however, T1 IIR-SNN and ReLU activated ANN both infer in single shot, difference being the neuronal functionality and output precision. This difference reduces further with ANNs using binary activations (ANNb) which also have 0/1 output, like SNNs. Thus, it is interesting to compare the performance of T1 IIR-SNN and ANNb which is the focus of Table \ref{annb}. We observe that IIR-SNN outperforms other ANNb approaches in most cases; our results are slightly lower on CIFAR100 compared to \cite{optimal}, however in their case, the ANN ReLU output is thresholded to saturate at 2, but the values are still analog. Note, in all cases, weights are full precision.

\vspace{-1mm}
\textbf{Efficacy of IIR-SNN with shallow networks.} As mentioned in section 4, gradual training is required due to spike vanishing at later layers for deep SNNs if direct transition from T5 to T1 is attempted. However, for shallow networks, if there is some propagation of spikes till the end, training with T1 following direct conversion from ANN might be possible. To validate this, we experiment with a VGG6 on CIFAR10 and the results are shown in Table \ref{tab4}. In this table, VGG6g denotes networks trained with gradual latency reduction and VGG6d denotes networks directly converted from ANN and trained using that particular timestep, so the result in the last row of Table \ref{tab4} is obtained by converting an ANN to SNN and directly training with 1 timestep. We observe that proposed gradual training scheme provides slightly higher accuracy compared to direct training in case of shallower networks, though it increases the training overhead. This is consistent with ANN domain results \cite{furlanello2018born,han2015}, where the authors achieve better performing networks using sequential model compression compared to direct compression.

\vspace{-1mm}
\textbf{IIR-SNN with and without batch-norm.} Recent works have achieved low latency by adopting batch-normalization (bn) suitably in SNNs \cite{zheng2021going,kim2020revisiting}. To disentangle the effect of bn from the proposed gradual latency reduction scheme and ensure that achieving convergence with 1 timestep is orthogonal to using bn, we perform ablation studies as shown in Table \ref{tab5}. For both CIFAR10 and CIFAR100, IIR-SNN enables training with T5 to T1 irrespective of using bn during ANN training. Using bn enhances accuracy, but sequential latency reduction can be performed independently from it. Also, \cite{zheng2021going} report that using threshold-dependent bn allows reducing latency up to a minimum of 6 timesteps, but IIR-SNN can go up to 1. Note, in IIR-SNN, bn is used only during ANN training and the bn parameters are fused with the weights during ANN-SNN conversion as proposed in \cite{rueckauer2017conversion}; bn is not used in SNN domain training, so the activations at each layer output remain binary.

\begin{table}[t]
\renewcommand{\arraystretch}{1.06}
\begin{minipage}[t]{.27\textwidth}
\begin{flushleft}
\vspace{-10mm}
\centering
\caption{Effect of IIR on VGG6 with CIFAR10}
\label{tab4}
\renewcommand{\arraystretch}{1.15}

\setlength{\tabcolsep}{5pt}

       \begin{tabularx}{\textwidth}{ccc}

\hline
T & VGG6g(\%) & VGG6d(\%) \\
\hline
5 & 90.61 & 90.15 \\
\hline

4 & 90.52 & 90.08 \\
\hline

3 & 90.40 & 89.91\\
\hline

2 & 90.05 &89.35 \\
\hline

1 & 89.10 & 88.64 \\
\hline

\vspace{-11mm}
\end{tabularx}

\end{flushleft}

\end{minipage}
\hspace{.5em}
\begin{minipage}[t]{.7\textwidth}
\renewcommand{\arraystretch}{0.9}
\begin{flushright}
\vspace{-10mm}
\centering
\caption{Accuracy(\%) of IIR-SNN with VGG16, D\textunderscore bn and D\textunderscore nbn denote dataset D with and without batch-norm respectively}
\label{tab5}
\renewcommand{\arraystretch}{1.15}

\setlength{\tabcolsep}{10pt}

       \begin{tabularx}{\textwidth}{ccccc}

\hline
T & CIFAR10\textunderscore bn & CIFAR10\textunderscore nbn & CIFAR100\textunderscore bn & CIFAR100\textunderscore nbn\\
\hline

5 & 93.90 & 92.15 & 71.58 & 69.86\\
\hline

4 & 93.87 & 91.95 &71.51 & 69.84\\
\hline

3 & 93.85 & 91.90 &71.46& 69.76\\
\hline

2 & 93.72 & 91.88 &71.43& 69.30\\
\hline

1 & 93.05 & 91.03 & 70.15 & 67.76\\
\hline

\vspace{-11mm}
\end{tabularx}
\end{flushright}
\end{minipage}
\end{table}

\vspace{-1mm}
\textbf{Skipping intermediate latency reduction steps.} Since the proposed IIR-SNN scheme increases training overhead due to sequential retraining, it is worth investigating if this cost can be reduced by skipping in between latency reduction steps. To that effect, we experiment with 2 cases- (i) training T5 followed by T3\textunderscore 5, followed by T1\textunderscore 3, (ii) IIR-SNN including all intermediate latencies with the sequence- T5, T4\textunderscore 5, T3\textunderscore 4, T2\textunderscore 3 and  T1\textunderscore 2. Interestingly, both these cases perform  comparably; for CIFAR10, we obtain 93.01\% and 93.05\% accuracy, respectively with 1 timestep for cases (i) and (ii). Again, these values for CIFAR100 are 69.92\% and 70.15\%, respectively, indicating  if the end goal is obtaining T1 SNN, training overhead can be reduced by skipping intermediate steps. However, in this paper, we report the whole spectrum of accuracy-latency trade-off from T5 to T1.

\vspace{-1mm}
\textbf{Is training following direct conversion from ANN with T1 thresholds feasible?} As mentioned above, a limitation of IIR-SNN is extra training overhead. Though it does not affect our primary goal (inference efficiency) , it would be better if direct conversion from ANN to T1 would be feasible. This is challenging due to layerwise spike decay as discussed in detail in section 4. In this part, we  revisit this from a different angle. First, we obtain the layerwise V$_{\text{th}}$ for T1 using IIR-SNN. Then, using this set of V$_{\text{th}}$, we convert a VGG16 ANN to SNN, on CIFAR10 and CIFAR100 to investigate whether these thresholds trained for T1 can enable direct training with T1 following ANN-SNN conversion. However, training failure occurs in this case too, since during IIR-SNN T1 training, the weights (w) get modified in addition to the thresholds, and layerwise spike propagation depends on both V$_{\text{th}}$ and w. Even if training converged in this case, the challenge would remain how to obtain the suitable V$_{\text{th}}$ for T1 without IIR, however it would indicate the existence of deep T1 SNNs which can be trained by ANN-SNN conversion followed by direct SNN domain T1 training. But in our experiments, we are unable to find such networks, further validating the need of IIR method.

\vspace{-1mm}
\textbf{IIR-SNN with magnitude based weight pruning.} With T1 IIR-SNN, we can obtain maximum compression possible in the temporal domain of SNNs, as 1 forward pass is the minimum for any network to produce output. Additionally, we hypothesize that T1 networks might be amenable to spatial pruning as there is significant redundancy in the layerwise weights of DNNs \cite{han2015,deepcompresion}. To investigate this, we perform experiments on T1 with magnitude based weight pruning \cite{deepcompresion}. Our result indicate that we can remove 90\% of the total weights of a VGG16 without large drop in performance. Using VGG16 (T1), we obtain 91.15\% accuracy on CIFAR10 and 68.20\% accuracy on CIFAR100, while retaining just 10\% of the spatial connections. This provides evidence that spatial pruning techniques can be combined with T1 IIR-SNNs to achieve holistic spatio-temporal compression, albeit with a slight hit to classification accuracy.
\begin{figure}[t]
\vspace{4mm}
  \centering
   \includegraphics[width=\linewidth]{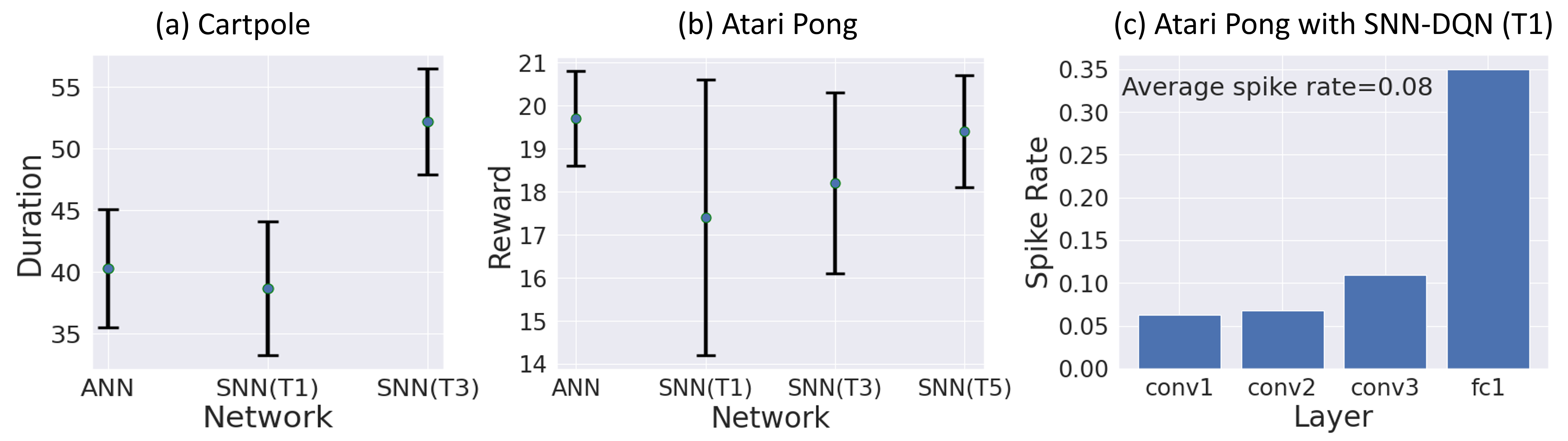}
  
   \caption{Average reward (errorbars depict mean$\pm$std),  using DQN with ANN and IIR-SNN- (a) on Cartpole and (b) Atari Pong, (c) layerwise spike rate with IIR-SNN DQN (T1) on Atari Pong.
   }
  \label{fig:fig3}
\vspace{-2em}
\end{figure}

\textbf{Reinforcement learning (RL) with IIR-SNN.} Due to their inherent recurrence, SNNs with multiple timesteps might be more useful in sequential decision-making (such as RL tasks) than static image classification. However, application of SNNs in RL may be limited if the latency is too high, since the agent has to make decisions in real-time. \cite{tan2020strategy} obtain high performing RL agents on Atari games with SNNs, but with 500 timesteps. Hence, in this section, we investigate if IIR-SNN technique can enable training SNNs for RL tasks with low latency. Experiments are performed using SNN based deep Q-networks (DQN) \cite{mnih2015human} on cartpole and Atari pong environments.  As shown in Fig.~\ref{fig:fig3}(a) and (b), for both cases, we can train SNNs up to 1 timestep. The rewards are obtained by averaging over 20 trials and plotted with errorbars showing mean$\pm$std. In (a), the reward is the duration of the cartpole remaining balanced; DQN with ANN, SNN(T1) and SNN(T3) achieve 40.3$\pm$4.8, 38.7$\pm$5.4, 52.2$\pm$4.3, respectively. While T1 SNN performs slightly worse compared to ANN-based DQN, SNN with T3 outperforms the ANN. Similar performance improvement over ANN using SNN for some RL tasks has been reported in \cite{tan2020strategy}. Next, we experiment with a more complex task, Atari pong game; in this case, T1 SNN achieves reward of 17.4$\pm$3.2 compared to ANN based DQN's 19.7$\pm$1.1. However, with T5 SNN, we obtain comparable reward (19.4$\pm$1.3) to ANN. Notably, with IIR-SNN, we can obtain SNNs for RL task (pong) with significantly lower latency compared to prior art. On pong, \cite{tan2020strategy} report reward of 19.8$\pm$1.3 using 500 timesteps, whereas, we obtain 19.4$\pm$1.3 reward using just 5 timesteps. So, IIR-SNN enables obtaining SNNs for RL tasks with comparable performance to ANNs with $\sim5$ timesteps, and up to 1 timestep with slightly lower performance. Moreover, such latency reduction translates to considerable energy savings which is critical for agents operating in the real world. The layerwise spike rates for SNN-DQN (T1) are shown in Fig.~\ref{fig:fig3}(c), the average spike rate is 0.08 only in this case, which results in 7.55X higher energy efficiency compared to ANN-DQN. Again, if we compare iso-performing networks to ANN-DQN, SNN-DQN (T5) infers with average spike rate  0.42, thus providing 5.22X higher energy efficiency compared to ANN-DQN.
The details of network topologies used, training setup and additional results are given in appendix \ref{rl-extra}.

\vspace{-3mm}
\section{Conclusion}
\label{conclusion}
\vspace{-3mm}
Bio-plausible SNNs hold promise as energy efficient alternatives to ANNs. However, to make SNNs suitable for edge deployment, mitigating high inference latency is critical. To that end, we propose IIR-SNN, a sequential SNN training approach which enables SNNs to infer in a single shot while maintaining comparable performance to their ANN counterparts. We first train an ANN followed by ANN-SNN conversion. Then SNN domain training is performed with 5 timesteps using direct input encoding. At each subsequent stage of training, latency is reduced gradually using the network trained at previous stage as initialization, this allows us to obtain deep SNNs up to unit latency. 
The proposed scheme enables inference with 25-33X higher energy efficiency compared to an equivalent ANN with similar accuracy. Furthermore, in comparison with other state-of-the-art SNNs, T1 IIR-SNNs reduce latency by 5-2500X without significant degradation of performance. Again, T1 SNNs get rid of the memory access overhead associated with fetching membrane potentials which is prevalent in SNNs with multiple timesteps. Moreover, inference with unit latency improves compute efficiency of SNNs too since the total spike count is reduced significantly compared to SNNs with high latency. We also investigate the efficacy of IIR-SNN in obtaining low latency SNN solutions for RL tasks (cartpole and Atari pong) and observe that SNN based DQNs can perform reasonably well even with just 1 timestep, and comparably to ANN based DQNs with 5 timesteps. In short, the proposed method tries to address the high inference latency issue of SNNs and paves the way of obtaining SNNs with lowest possible latency (1), albeit while incurring some extra training overhead and a slight accuracy drop in some cases.

\section{Acknowledgements}
\label{acknowledgements}

This work was supported in part by the Center for Brain Inspired Computing (C-BRIC), one of the six centers in JUMP, a Semiconductor Research Corporation (SRC) program sponsored by DARPA, by the SRC, the National Science Foundation, Intel Corporation, the DoD Vannevar Bush Fellowship, and by the U.S. Army Research Laboratory.

\bibliography{main.bib}
\bibliographystyle{unsrt} 
\newpage

\appendix
\section{Appendix}
\subsection{Training Methodology}
\label{app1}
\subsubsection{surrogate-gradient based learning}
To train deep SNNs, surrogate-gradient based backpropagation through time (BPTT) \cite{werbos1990backpropagation} is used to perform temporal as well as  spatial error credit assignment. Spatial credit assignment is achieved by spatial error distribution  across layers, while the network is unrolled in time for temporal credit assignment. The output layer neuronal dynamics is governed by-
\begin{equation}\label{eqn:4}
u_i^t=u_i^{t-1}+\sum_{j} w_{ij}o_j^t,
\end{equation}
here $u_i$ corresponds to the membrane potential of i-th neuron of final (L-th) layer. The final layer neurons just accumulate the potential over time for classification purpose, without emitting spikes. We pass these accumulated final layer outputs through a softmax layer which gives the class-wise probability
distribution and then the loss is calculated using the cross-entropy between the true output and the
network’s predicted distribution. The governing equations are-
\begin{equation}\label{eqn:5}
Loss=-\sum_{i} y_{i}log(s_i),
\end{equation}
\begin{equation}\label{eqn:6}
s_i=\frac{e^{u_i^T}}{\sum_{k=1}^{N} e^{u_k^T}},
\end{equation}
where Loss represents the loss function, y denotes true label, s is the predicted label, T is the total number of timesteps and
N is the number of classes. The derivative of the loss w.r.t. the membrane potential of the neurons in the final layer is-
\begin{equation}\label{eqn:7}
\frac{\partial Loss}{\partial \bm{u}_L^T}=\bm{s}-\bm{y},
\end{equation}
and the weight updates at the output layer are done as-
\begin{equation}\label{eqn:8}
w_{ij,L}=w_{ij,L}-\eta\Delta w_{ij,L},
\end{equation}
\begin{equation}\label{eqn:9}
\Delta w_{ij,L}=\sum_{t} \frac{\partial Loss}{\partial w_{ij,L}^t}=\sum_{t} \frac{\partial Loss}{\partial \bm{u}_{L}^T} \frac{\partial \bm{u}_{L}^T}{\partial w_{ij,L}^t}=\frac{\partial Loss}{\partial \bm{u}_{L}^T} \sum_{t}  \frac{\partial \bm{u}_{L}^T}{\partial w_{ij,L}^t},
\end{equation}
where $\eta$ is the learning rate, and $w_{ij,L}^t$ denotes the weight between i-th neuron at layer $L$ and j-th neuron at layer $L-1$ at timestep t. Output layer neurons are non-spiking, hence the non-differentiability issue is not involved here. The hidden layer parameter update is given by-
\begin{equation}\label{eqn:10}
\Delta w_{ij,k}=\sum_{t} \frac{\partial Loss}{\partial w_{ij,k}^t}=\sum_{t} \frac{\partial Loss}{\partial o_{i,k}^t} \frac{\partial o_{i,k}^t}{\partial u_{i,k}^t}\frac{\partial u_{i,k}^t}{\partial w_{ij,k}^t},~~~k=2,3,... L-1
\end{equation}
where $o_{i,k}^t$
is the spike-generating function (Eqn.~3), k is layer index. We approximate the gradient of this function w.r.t. its input using the linear surrogate-gradient \cite{bellec2018long} as-
\begin{equation}\label{eqn:11}
\frac{\partial o}{\partial u}= \gamma max\{0,1-|\frac{u-v}{v}|\},
\end{equation}
where $\gamma$ is a hyperparameter chosen as 0.3 in this work. Again, the layerwise threshold is updated using-
\begin{equation}\label{eqn:12}
v_{l}=v_{l}-\eta\Delta v_{l},
\end{equation}
\begin{equation}\label{eqn:13}
\Delta v_{l}=\sum_{t} \frac{\partial Loss}{\partial v_{l}}=\sum_{t} \frac{\partial Loss}{\partial \bm{o}_l^t}\frac{\partial \bm{o}_l^t}{\partial \bm{z}_l^t}\frac{\partial \bm{z}_l^t}{\partial v_l}=\sum_{t} \frac{\partial Loss}{\partial \bm{o}_l^t}\frac{\partial \bm{o}_l^t}{\partial \bm{z}_l^t}\left(\frac{-v_l\bm{o}_l^{t-1}-\bm{u}_l^t}{v_l^2}\right)
\end{equation}

and during training stages where more than one timestep is involved, the leak is updated as-
\begin{equation}\label{eqn:14}
\lambda_{l}=\lambda_{l}-\eta\Delta \lambda_{l},~~~\text{and}~~~\Delta \lambda_{l}=\sum_{t} \frac{\partial Loss}{\partial \lambda_{l}}=\sum_{t} \frac{\partial Loss}{\partial \bm{o}_l^t}\frac{\partial \bm{o}_l^t}{\partial \bm{u}_l^t}\frac{\partial \bm{u}_l^t}{\partial \lambda_l}=\sum_{t} \frac{\partial Loss}{\partial \bm{o}_l^t}\frac{\partial \bm{o}_l^t}{\partial \bm{u}_l^t}\bm{u}_l^{t-1}
\end{equation}

\begin{algorithm}[H]
   \caption{Procedure of spike-based backpropagation learning for an iteration.}
   \label{alg:surrogate}
\begin{algorithmic}
   \STATE {\bfseries Input:} pixel-based mini-batch of input ($\bm{X}$) - target ($\bm{Y}$) pairs, total number of timesteps ($T$), number of layers ($L$), pre-trained ANN weights ($\bm{W}$), membrane potential ($\bm{U}$), layer-wise membrane leak constants ($\bm{\lambda}$), layer-wise firing thresholds ($\bm{V}$), learning rate ($\eta$)
   \STATE {\bfseries Initialize:} $\bm{U}_l^t = 0,~\forall l = 1,...,L$
   \STATE //~{\bfseries Forward Phase}
   \FOR{$t \leftarrow 1$ {\bfseries to} $T$}
          
        \STATE $\bm{O}_1^t = \bm{X};$ 
        \FOR{$l \leftarrow 2$ {\bfseries to} $L-1$}         
            \STATE //~membrane potential integrates weighted sum of spike-inputs
            \STATE $\bm{U}_l^t~= \lambda_l\bm{U}_l^{t-1} + \bm{W}_{l}* \bm{O}_{l-1}^t$
        \IF {$\bm{U}_l^t > V_l$}
        \STATE //~if membrane potential exceeds $V_l$, a neuron fires a spike
            \STATE $\bm{O}_{l}^t = 1,~\bm{U}_l^t = \bm{U}_l^t-V_l$
        \ELSE
        \STATE //~else, output is zero
            \STATE $\bm{O}_{l}^t  = 0$
        \ENDIF
        \ENDFOR    
        \STATE //~final layer neurons do not fire
        \STATE $\bm{U}_{L}^t~= \bm{U}_{L}^{t-1} + \bm{W}_{L}* \bm{O}_{L-1}^t$
   \ENDFOR
   \STATE //calculate loss, Loss=cross-entropy($\bm{U}_{L}^T,\bm{Y}$)
   \STATE //~{\bfseries Backward Phase}
   \FOR{$t \leftarrow T$ {\bfseries to} $1$}
       \FOR{$l \leftarrow L-1$ {\bfseries to} $1$}    
       \STATE //~evaluate partial derivatives of loss with respect to the trainable parameters by unrolling the network over time
       \STATE $\triangle \bm{W}_l^t = \frac{\partial Loss}{\partial \bm{O}_l^t}\frac{\partial \bm{O}_l^t}{\partial \bm{U}_l^t}\frac{\partial \bm{U}_l^t}{\partial \bm{W}_l^t},~~
       \Delta V_{l}^t= \frac{\partial Loss}{\partial \bm{O}_l^t}\frac{\partial \bm{O}_l^t}{\partial \bm{z}_l^t}\frac{\partial \bm{z}_l^t}{\partial V_l},~~\Delta\lambda_{l}^t= \frac{\partial Loss}{\partial \bm{o}_l^t}\frac{\partial \bm{o}_l^t}{\partial \bm{u}_l^t}\frac{\partial \bm{u}_l^t}{\partial \lambda_l}$
       \ENDFOR
   \ENDFOR
   \STATE //update the parameters
   $\bm{W}_{l}=\bm{W}_{l}-\eta\sum_{t}\Delta \bm{W}_{l}^t, V_{l}=V_{l}-\eta\sum_{t}\Delta V_{l}^t,\lambda_{l}=\lambda_{l}-\eta\sum_{t}\Delta \lambda_{l}^t$
   \end{algorithmic}
\end{algorithm}

\subsection{Experimental details}
\label{app2}

\subsubsection{Network architecture}
\vspace{-1mm}
We modify the VGG and ResNet architectures slightly to facilitate ANN-SNN conversion. 3 plain convolutional layers of 64 filters are appended after the input
layer in the ResNet architecture \cite{sengupta2019going}. In all cases, average
pooling (2$\times$2) is used and the basic block in ResNet employs a stride of 2 when the
number of filters increases. 1$\times$1 convolution is used in the shortcut path of basic blocks where the number of
filters is different in input and output. The architectural details are-

VGG6: \{64, A, 128, 128, A\}, Linear

VGG16: \{64, D, 64, A, 128, D, 128, A, 256, D, 256, D, 256, A, 512, D, 512, D, 512, A, 512, D, 512,
D, 512\}, Linear

ResNet20: \{64, D, 64, D, 64, A, 64BB, 64BB, 128BB ($/$2), 128BB, 256BB ($/$2), 256BB, 512BB
($/$2), 512BB\}

BB: basic block, Linear: \{4096, D, 4096, D, number of classes\}, D: Dropout (probability (p)-ANN: 0.5, SNN: 0.2),
A: Average Pooling (kernel size = 2$\times$ 2)
\vspace{-1mm}
\subsubsection{Training Hyperparameters}
\vspace{-1mm}
Standard data augmentation
techniques are applied for image datasets such as padding by 4 pixels on each side, and $32\times32$ cropping by randomly sampling from the padded image or its horizontally flipped version (with 0.5 probability of flipping). The original $32\times32$ images are used during testing.  Both training and testing data are normalized using channel-wise mean and standard deviation calculated from training set. The ANNs are trained with cross-entropy loss with stochastic gradient descent optimization (weight decay=0.0005, mometum=0.9). The ANNs domain are trained for 500 and 90 epochs for CIFAR and ImageNet respectively, with an initial learning rate of 0.01. The learning rate 
is divided by 5 at epochs of 0.45, 0.7 and 0.9 fraction of total epochs. The ANNs are trained with batch-norm (bn) and the bn parameters are fused with the layerwise weights during ANN-SNN conversion following \cite{rueckauer2017conversion}, we do not have bias terms as bn is used. Additionally, dropout \cite{srivastava2014dropout} is used as the regularizer with a constant
dropout mask with dropout probability=0.5 across all timesteps
while training the SNNs. Again, since max-pooling causes significant
information loss in SNNs \cite{diehl2015fast}, we use average-pooling
layers to reduce the feature maps. 
  The ANN, the weights are initialized using He initialization \cite{he2015delving}. Upon conversion, at training iteration with the timestep fixed, the SNNs are trained for 300 epochs with cross-entropy loss
and adam optimizer (weight decay=0). Initial learning rate is chosen as 0.0001,
which is divided by 5 at epochs of 0.6, 0.8 and 0.9 fraction of total epochs. While training the SNNs, dropout probability is kept at 0.2.

\vspace{-2mm}
\subsection{Comparison of accuracy versus latency trade-off}
\label{app3-1}
\begin{figure}[t]
  \centering
\includegraphics[width=\linewidth]{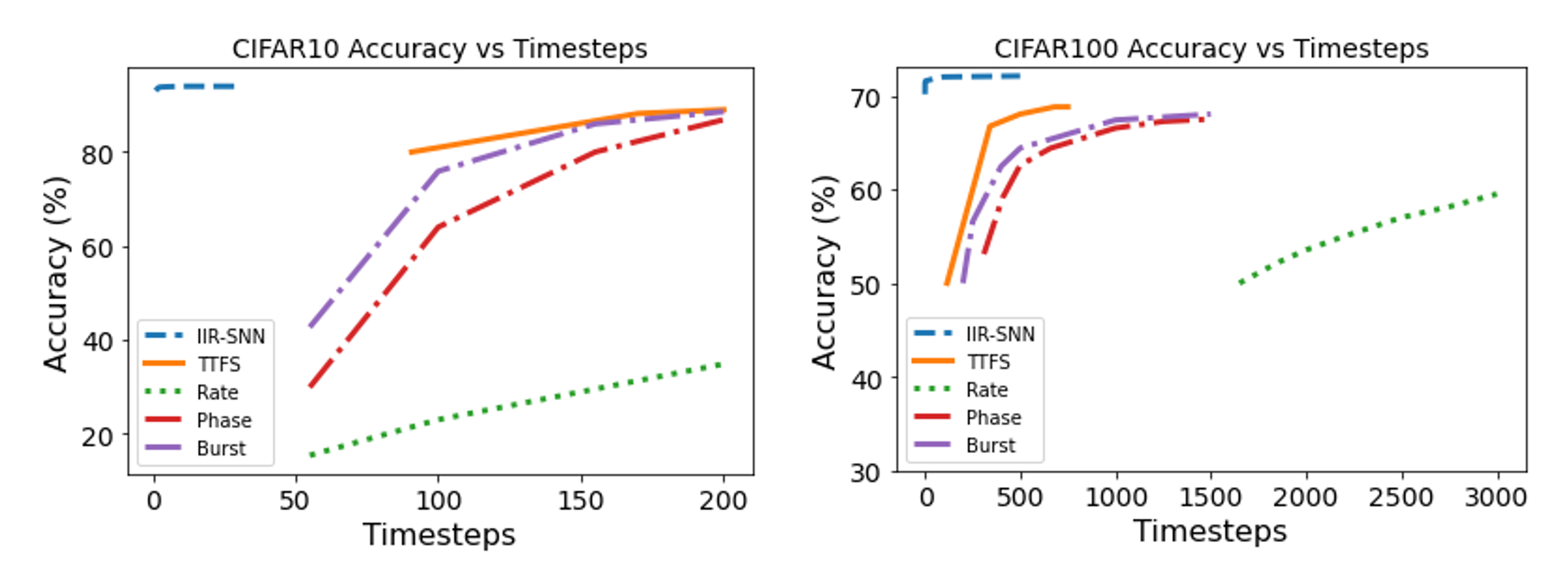}
   \caption {Accuracy versus latency curve for various SNN coding methods on VGG16, the values for TTFS \cite{park2020t2fsnn}, phase \cite{kim2018deep}, burst \cite{park2019fast} and rate \cite{rueckauer2017conversion} have been adopted from \cite{park2020t2fsnn}.}.
   \vspace{-2em}
  \label{fig:compare_Park}
\end{figure}
\vspace{-1mm}
In this part, we compare our results with other SNN encoding schemes in terms of timesteps required to reach convergence for a VGG16 network and the results are shown in Fig.~\ref{fig:compare_Park}. The figure demonstrates the results of “T2FSNN” encoding, where the timing of spikes carries information and other rate and temporal coding schemes including “Rate” \cite{rueckauer2017conversion}, “Phase” \cite{kim2018deep}, and “Burst” coding \cite{park2019fast}. The left plot in Fig.~\ref{fig:compare_Park} shows $\sim200$ timesteps is needed for the fastest convergence among these encoding methods for CIFAR10, whereas, we achieve $\sim 93.05\%$ accuracy in just 1 timestep. Similarly, IIR-SNN obtains 70.15\% accuracy in 1 timestep with VGG16 on CIFAR100. The graph on the right in Fig.~\ref{fig:compare_Park} shows the results for VGG16 on CIFAR100 using “T2FSNN”, “Burst”, “Phase” and “Rate”. As can be seen, “T2FSNN” reaches 68\% roughly at 500 steps, “Burst” at 1500, “Phase” at 2000, and “Rate” fails to cross 60\% even at 3000 timesteps.  Notably, IIR-SNN not only is able to reduce latency by 2 to 3 orders of magnitude compared to these works, but also outperforms them in top-1 accuracy by $\sim2\%$ using unit inference latency. So, IIR-SNN enhances the performance on both ends of the accuracy-latency trade-off.

\subsection{Computational Cost}
\label{app3-2}
\vspace{-1mm}
The number of operations in an ANN layer is given as-
\vspace{-2mm}
\begin{equation*}
\# \text{ANN}_{\text{ops}} = \left\{
                \begin{array}{ll}
                  k_w\times k_h\times c_{in}\times h_{out} \times w_{out} \times c_{out},~~~\text{Conv layer}\\
                  n_{in}\times n_{out},~~~~~~~~~~~~~~~~~~~~~~~~~~~~~~~~~~~~~~~~~~~~~\text{Linear layer}
                \end{array}
              \right.
\end{equation*}

where $k_w (k_h)$ denote filter width (height), $c_{in} (c_{out})$ is number of input (output) channels,
$h_{out} (w_{out})$ is the height (width) of the output feature map, and $n_{in} (n_{out})$ is the number of input
(output) nodes.

\subsection{Additional Results on Reinforcement learning}
\label{rl-extra}
\begin{figure}[t]
\vspace{-8mm}
  \centering
   \includegraphics[width=\linewidth]{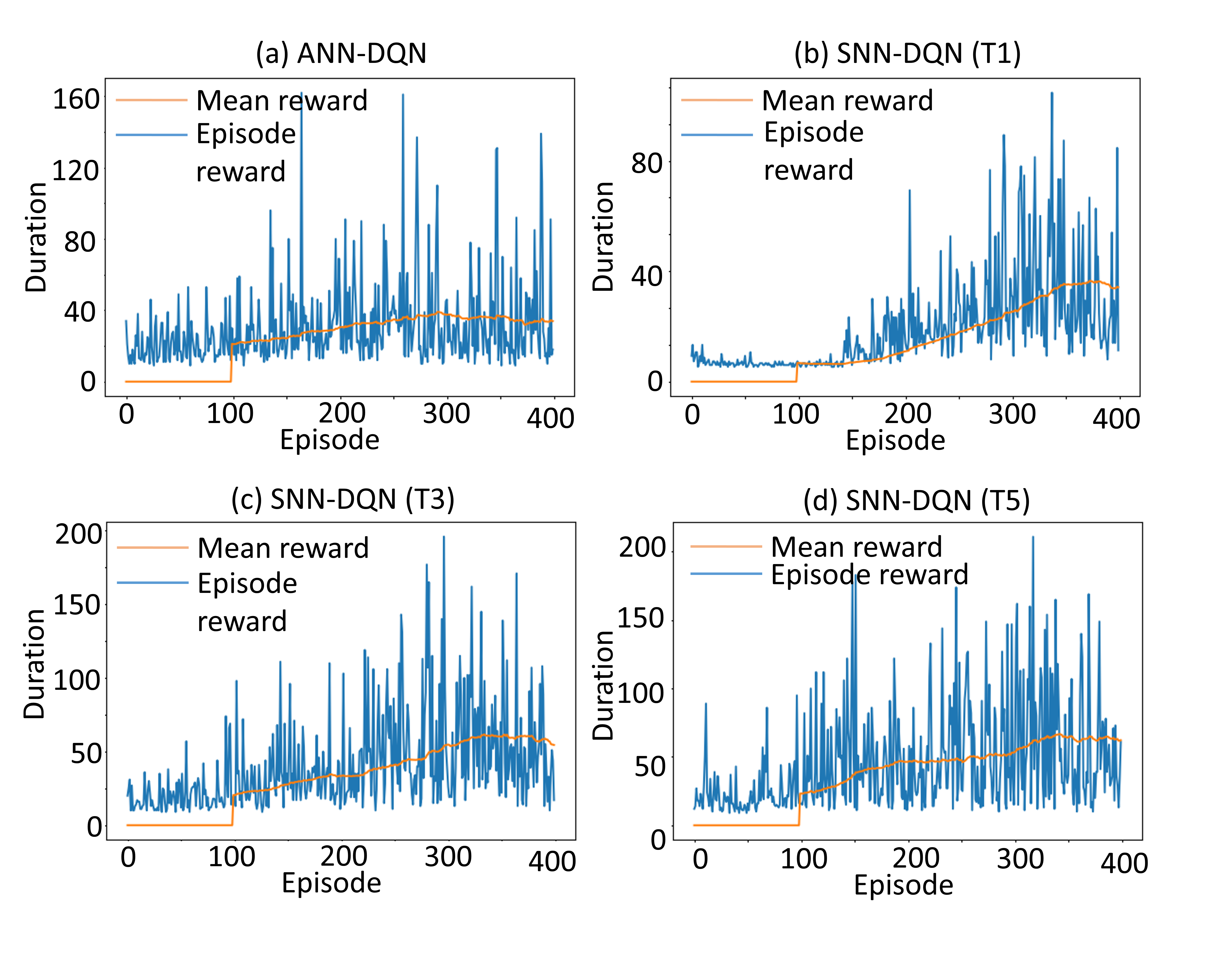}
  
   \caption{Rewards during training on Cartpole environment with- (a) ANN-DQN and (b) SNN-DQN (T1), (c) SNN-DQN (T3), and (d) SNN-DQN (T5).
   }
  \label{fig:cartpole-all}
\vspace{-1em}
\end{figure}

\begin{figure}[t]
\vspace{-12mm}
  \centering
   \includegraphics[width=\linewidth]{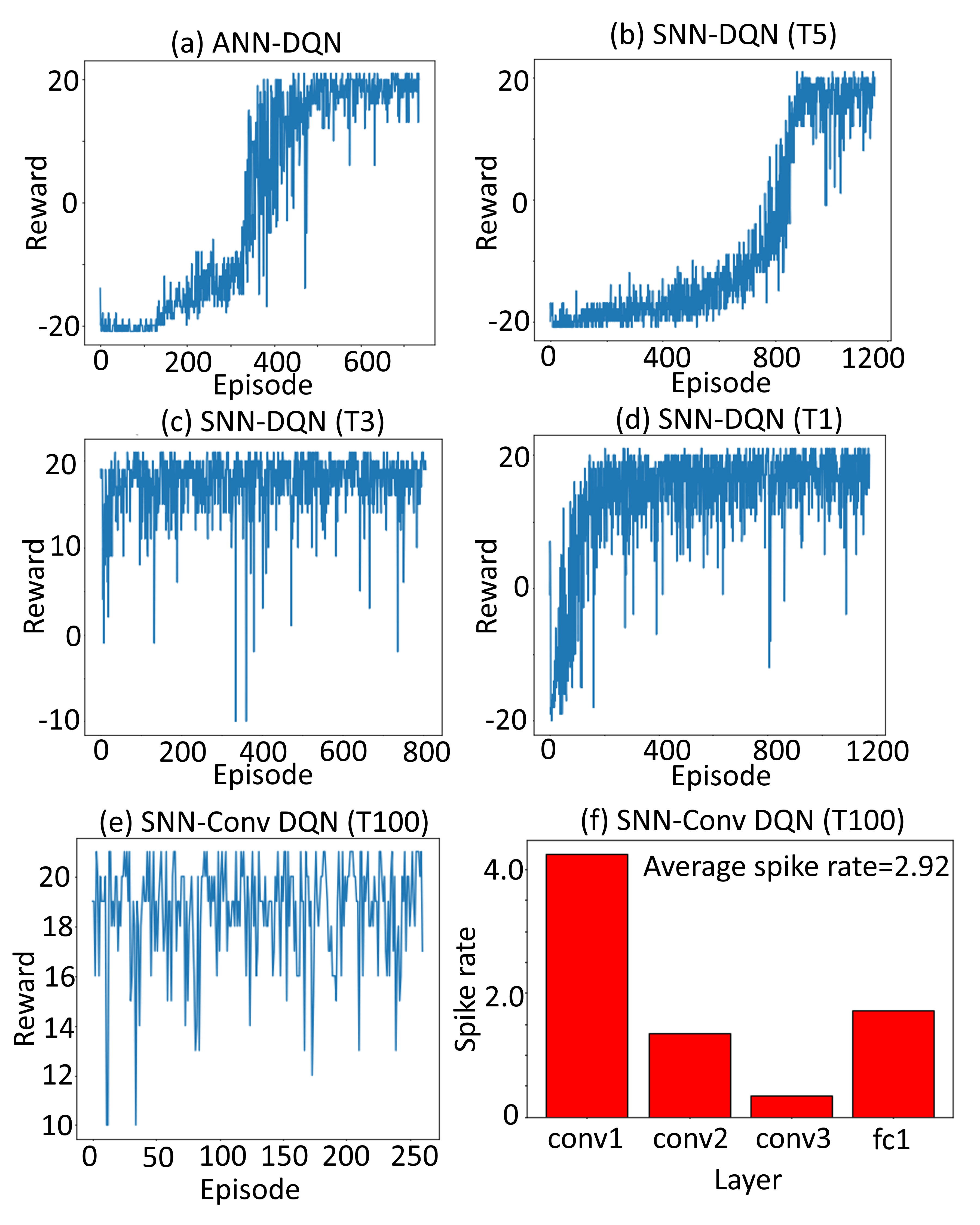}
  
   \caption{Rewards during training on Atari pong environment with- (a) ANN-DQN and (b) SNN-DQN (T5), (c) SNN-DQN (T3), and (d) SNN-DQN (T1), (e) SNN-Conv DQN (T100), and (f) layerwise spike rate for the network in (e).
   }
  \label{fig:pong-all}
\vspace{-1em}
\end{figure}

In this section, we provide supplementary results on our experiments with IIR-SNN on RL tasks (Cartpole and Atari pong). The details of network architectures, training hyperparameters and training dynamics are described in the following.
\vspace{-2mm}
\subsubsection{Results on Cartpole}
\vspace{-2mm}
For the cartpole task, we use a small network with 3 convolutional and a fully connected layer. The first convolutional layer has 16 5X5 filters with stride 2,  second and third convolutional layers have 32 5X5 filters with stride 2 each. The fully connected layer has 2 neurons, corresponding to the number of actions. We use RMSProp as optimizer for training. The networks are trained with batch size of 128, discount factor ($\gamma$) of 0.999 and replay memory of 10000. Figure \ref{fig:cartpole-all} shows the results for the cartpole environment. We repeat all experiments for 400 episodes, in this game, the duration for which the agent is able to continue the game (keep the balance), is the reward; so higher game duration means better performance. The blue trajectory shows the rewards at each episode, which has some variation across episodes, the yellow curve shows the accumulated average reward. As can be seen, the SNN outperforms the ANN in terms of reward with just 3 timesteps. For SNN-DQN with 1 timestep (T1), the SNNs performs slightly worse compared to ANN-DQN. As we increase the simulation time-window (number of timesteps), the SNN performance improves further as expected, with the average reward for T5 being 52.2$\pm$4.3 (mean$\pm$standard deviation), whereas for the ANN, this metric is 40.3$\pm$4.8. We anticipate this improvement in multi-timestep SNNs is due to inherent memory of previous inputs which results from the residual membrane potential in the spiking neurons. As a result, SNNs offer an advantage compared to feed-forward ANNs for tasks with a sequential component. In RL tasks, since the decision-making is sequential and past frames possess some information about what the next plausible state and corresponding action could be, SNNs might be better suited to leverage such kind of environment. However, cartpole balancing is a very simple task, so to further explore the potential of the proposed IIR-SNN in obtaining low latency solutions for SNN-based RL, we next apply the IIR SNN-DQN framework to Atari pong environment.

\subsubsection{Results on Atari pong}
Atari pong is a two-dimensional gym environment that simulates table tennis. Here, the agent steps through the environment by observing frames of the game (reshaped to 84X84 pixels), interacts with the environment with 6 possible actions, and receives feedback in the form of the change in the game score. 
For our experiments, we first train an ANN based deep Q network (DQN), where we use the same DQN proposed in \cite{mnih2015human} with 3 convolution layers and 2 fully connected layers. The first convolution layer has 32 8X8 filters with stride 4. The second convolution layer has 64 4X4 with stride 2. The third convolution layer has 64 3X3 filters with stride 1. The fully connected layer has 512 neurons. The number of neurons in the final layer for any game depends on the number of valid actions for that game, which in this case is 6. Training is performed with Adam optimizer with learning rate 0.00001, batch size 128, discount factor($\gamma$) 0.99 and replay buffer size 100000. Results of our experiments are shown in Fig.~\ref{fig:pong-all}. Rewards obtained during ANN-DQN training are depicted in Fig.~\ref{fig:pong-all} (a). As mentioned in the discussion of RL-SNN in section 5 of the main manuscript, ANN-DQN achieves reward of 19.7$\pm$1.1. The training dynamics using IIR-SNN DQN with T5, T3 and T1 are shown in Fig.~\ref{fig:pong-all} (b), (c), (d) respectively. Reward and efficiency analysis of these networks compared to ANN-DQN is given in section 5 of the main manuscript. Additionally, here we compare the performance of IIR-SNN to converted SNNs as shown in Fig.~\ref{fig:pong-all} (e). Note, these SNNs do not undergo any training in SNN domain, rather they are converted from corresponding ANN-DQNs and used for inference. Reward obtained the converted SNN with 100 timesteps is 19.4$\pm$1.3, so converted SNN-DQNs (SNN-conv DQNs) perform comparably to ANN-DQNs as also reported in \cite{tan2020strategy}, however the bottleneck is they require $\sim$100 timesteps for high performance. Using IIR-SNN, we obtain T5 SNN-DQNs with comparable performance to their ANN counterparts, but with considerably lower compute cost (5.22X). We have reported the spike rates for T1 SNN-DQN in Fig.~\ref{fig:fig3} (c). For comparison purposes, we also consider the layerwise spike rate for the converted SNN-DQN as shown in Fig.~\ref{fig:pong-all} (f),  of the main text. In this case, the average spike rate is 2.92, which leads to 1.75X higher energy efficiency for the SNN-conv DQN (T100) compared to ANN-DQN. As a result, IIR-SNN DQN (T5) provides 2.98X improvement in energy efficiency over SNN-conv DQNs as proposed in \cite{tan2020strategy}, while achieving comparable performance.

\end{document}